\documentclass[unnumsec,webpdf,contemporary,large]{oup-authoring-template}%

\graphicspath{{Fig/}}

\theoremstyle{thmstyleone}%
\theoremstyle{thmstyletwo}%
\theoremstyle{thmstylethree}%
\newtheorem{definition}{Definition}

\begin{document}

\journaltitle{Journal Title Here}
\DOI{DOI HERE}
\copyrightyear{2022}
\pubyear{2019}
\access{Advance Access Publication Date: Day Month Year}
\appnotes{Paper}

\firstpage{1}


\title[GiG Learning Framework for DTI Prediction]{A Graph-in-Graph Learning Framework for Drug-Target Interaction Prediction}

\author[1,$\ast$]{Yuehua Song}
\author[1]{Yong Gao}

\authormark{Y. Song and Y. Gao}


\address[1]{\orgdiv{Department of Computer Science}, \orgname{University of British Columbia Okanagan}, \orgaddress{\street{Kelowna}, \postcode{V1V 1V7}, \state{British Columbia}, \country{Canada}}}

\corresp[$\ast$]{Corresponding Author: Yuehua Song (\href{email:email-id.com}{yuehua@student.ubc.ca})}

\received{Date}{0}{Year}
\revised{Date}{0}{Year}
\accepted{Date}{0}{Year}



\abstract{Accurately predicting drug-target interactions (DTIs) is pivotal for advancing drug discovery and target validation techniques. While machine learning approaches including those that are based on Graph Neural Networks (GNN) have achieved notable success in DTI prediction, many of them have difficulties in effectively integrating the diverse features of drugs, targets and their interactions. To address this limitation, we introduce a novel framework to take advantage of the power of both transductive learning and inductive learning so that features at molecular level and drug-target interaction network level can be exploited. 
Within this framework is a GNN-based model called Graph-in-Graph (GiG) that represents graphs of drug and target molecular structures as meta-nodes in a drug-target interaction graph, enabling a detailed exploration of their intricate relationships. To evaluate the proposed model, we have compiled a special benchmark comprising drug SMILES, protein sequences, and their interaction data, which is interesting in its own right. Our experimental results demonstrate that the GiG model significantly outperforms existing approaches across all evaluation metrics, highlighting the benefits of integrating different learning paradigms and interaction data. 
}


\maketitle

\section{Introduction}
Accurate prediction of drug-target interactions (DTIs) plays a key role in modern drug development and precision medicine. The approval process for a new drug by the U.S. Food and Drug Administration (FDA) typically takes 10 to 17 years \cite{ashburn2004drug}, and any failure in the drug development process can lead to significant financial losses \cite{pereira2016boosting}. By understanding how drugs interact with their potential targets, researchers can design more effective drug molecules, accelerating the development process and reducing the risk of failure. Additionally, precision medicine depends on a thorough understanding of an individual's disease and therapeutic response, which includes how drugs interact with specific biomarkers. Improvements in DTI prediction can aid in creating personalized treatment plans, ensuring that patients receive the most suitable medications for their conditions, thereby maximizing therapeutic efficacy and minimizing adverse effects \cite{hopkins2008network}.

Since the experimental determination of compound-protein interactions or potential DTIs remains highly challenging \cite{haggarty2003multidimensional, kuruvilla2002dissecting}, there is a growing need for the development of effective in silico prediction methods. These predicted interactions can provide complementary and supporting evidence for experimental studies. Initially, DTI prediction methods involved docking simulations \cite{cheng2007structure, rarey1996fast}, but this approach was limited when applied to targets with unknown 3D structures. Over the past decades, it has been observed that existing interactions among drugs and targets provide a substantial amount of information that can be utilized to help predict DTIs effectively. These interaction data can be modeled by various interaction networks, such as drug-target networks and drug-disease networks. Early studies for DTI prediction focused on extracting local network structures from these network data to understand the relationship between specific pairs of drugs and targets, employing techniques and results from network science \cite{berger2009network, yildirim2007drug, nacher2008global, campillos2008drug}. 
In recent years, machine learning methods have gained significant attention in this area. Models based on matrix factorization (MF) \cite{zheng2013collaborative, luo2017network}, support vector machines (SVM), and random forests (RF) \cite{atas2023approach} have demonstrated competitive DTI prediction performance by leveraging extracted descriptors, which are numerical representations that encode key chemical and biological properties of drugs and targets.
A drawback of these traditional ML methods is that they require well-designed features to represent drugs and targets \cite{chen2018machine}. 
More recently, deep learning techniques have shown promise in predicting DTIs, including deep Random Walk, Convolutional Neural Networks (CNNs), and GNNs. These deep learning models learn embeddings of drugs and targets in the DTI networks and use these embeddings for DTI prediction, either directly or as a byproduct of reconstructing the interaction network. The underlying assumption, which is widely accepted, posits that drugs similar in the interaction networks are likely to interact with similar targets \cite{keiser2007relating}. Current efforts in this direction focus on using deep learning models to better integrate various interaction networks \cite{liu2021gadti, wang2014drug, wan2019neodti}. For instance, in NeoDTI, an autoencoder model is employed to predict DTIs on a heterogeneous graph that includes drug-drug, protein-protein, and drug-disease interaction networks. The embeddings generated by this model capture diverse relationships from the various interaction networks. Since learning occurs on interaction networks of a collection of drugs and targets, whether homogeneous or heterogeneous, these models are primarily categorized as transductive learning models. The information utilized by these models for DTI prediction is based on the network relationships between the drugs and targets and no molecular-level features or attributes are used.

\begin{figure}[!t]%
\centering
\includegraphics[width=1\columnwidth, keepaspectratio]{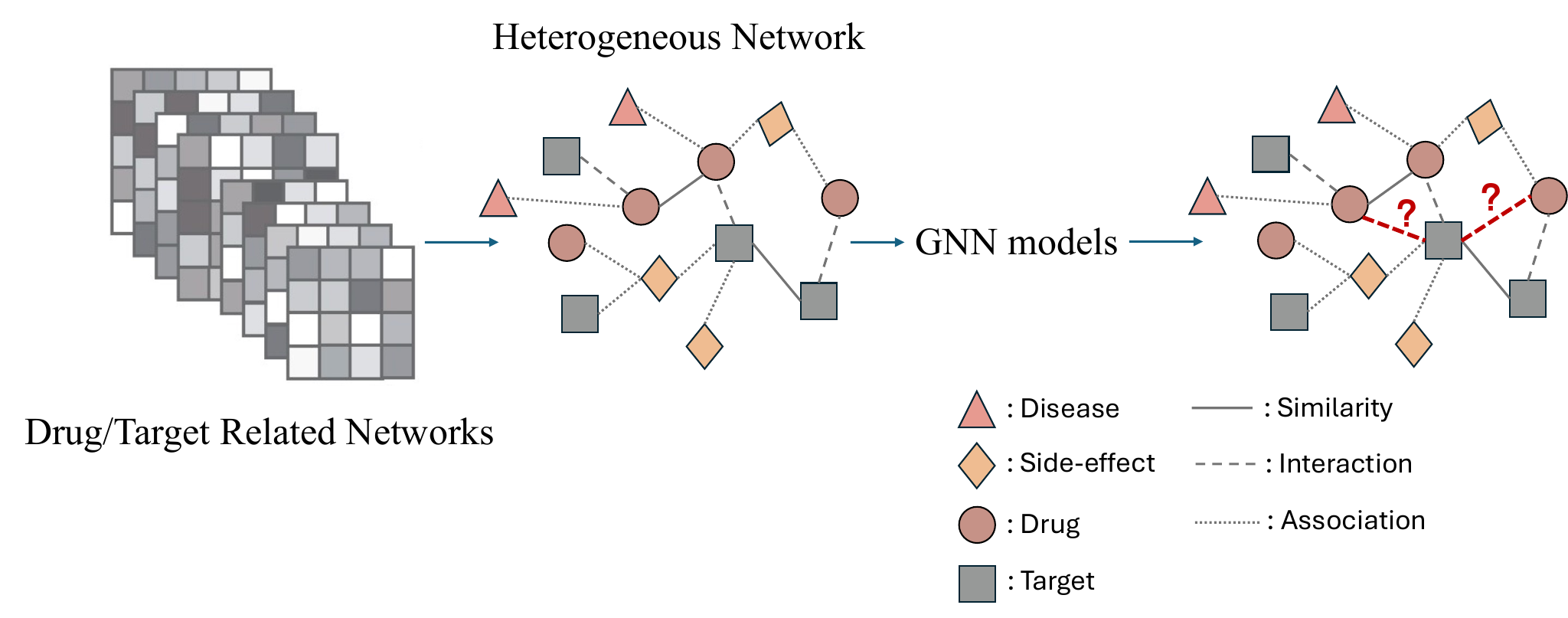}
\caption{The process for DTI task.}
    \label{fig:dti}
\end{figure}

While DTI prediction aims to identify potential interactions between a drug and a target, a related task known as Drug-Target Affinity (DTA) prediction, quantifies the strength of these interactions, specifically the binding affinity of the drug to the target \cite{nguyen2021graphdta, jiang2020drug, nguyen2021gefa, qi2024drug, jiang2022sequence}. These models are typically inductive learning models, where the training data for such models consist of pairs of drug molecular graphs and target structural graphs along with their affinity scores. 

\begin{figure}[!t]%
\centering
\includegraphics[width=1\columnwidth, keepaspectratio]{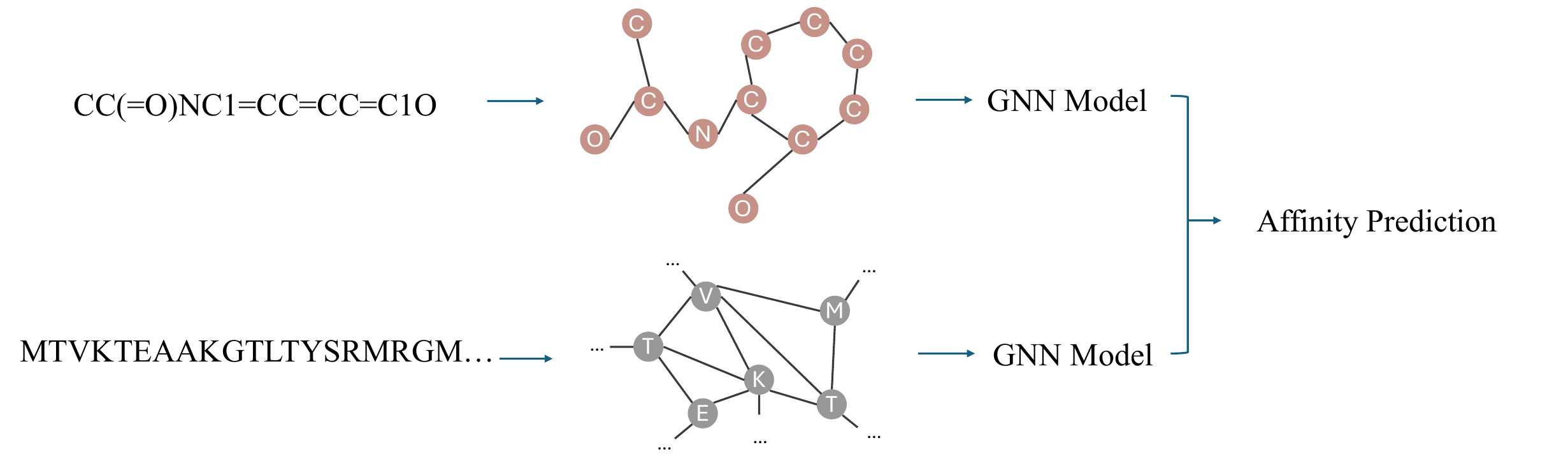}
\caption{The process for DTA task.}
    \label{fig:dta}
\end{figure}

As shown in Figure \ref{fig:dti}, common deep learning models for DTI prediction learn from drug-target interaction data presented as drug-target networks. The DTI task focuses on predicting drug-target relationships from a macro perspective, implying that predictions are made across an entire graph. The learning for such models is essentially a transductive learning problem, where new DTIs are predicted based on existing interactions within the DTI networks. Usually, no data on the physical or chemical attributes at the molecular level are directly used \cite{qu2024graph, liu2021gadti}. On the other hand, as illustrated in Figure \ref{fig:dta}, the typical approach for DTA prediction adopts a more microscopic perspective. Here, deep learning methods are used to predict the affinity between a drug and a target using the molecular graph of the drug and the structure graph of the target. 

In this work, we propose a GNN model for DTI prediction that integrates interaction network data with the molecular graphs of drugs and the sequence data of targets. Our model, named the Graph-in-Graph (GiG) model, is a hierarchical GNN that incorporates interactions among a collection of drugs and targets, alongside their molecular graphs and contact maps \cite{emerson2017protein} are used as features to enhance the interaction network data. By using the interaction information of a group of drugs and targets as well as individual features of the molecules, our model is a unique combination of transductive learning and inductive learning. To validate our proposed model, we have constructed a comprehensive dataset consisting of a DTI network, chemical expression of the drugs given using SMILES notation, and protein sequences. We believe this will also be a valuable resource for other researchers in this area. Using this dataset, extensive experiments have been conducted to evaluate our model. Our experimental results demonstrate that the GiG model significantly outperforms existing approaches across all evaluation metrics, highlighting the benefits of integrating different learning paradigms and interaction data.

\section{Materials and Methods}
\subsection{Our Datasets}
To validate our proposed model, we have constructed a comprehensive dataset, which we believe can be a valuable benchmark for this area. Our dataset consists of a DTI network on 708 drugs and 1512 targets, the chemical expressions for the molecular structures of the drugs, and the sequence data for the targets. The chemical expression for the molecule of a drug is represented by a ASCII string in SMILES (Simplified Molecular Input Line Entry System) -- a widely used notation system to describe the structure of chemical compounds \cite{weininger1988smiles}. The structures of the targets are represented by their contact map. Our dataset is unique in that it combines three types of data—chemical structures, protein sequences, and interaction networks—into a single dataset. To the best of our knowledge, it is the first benchmark with such combination in the domain of drug discovery. A summary of the statistics of the dataset is given in Table \ref{dataset_summary}. 

\begin{table}[!t]
\caption{Dataset Summary\label{dataset_summary}}%
\begin{tabular*}{\columnwidth}{@{\extracolsep\fill}ll@{\extracolsep\fill}}
\toprule
Statistic & Value  \\
\midrule
Number of Drugs        & 708   \\
Number of Targets      & 1512  \\
Number of Interactions & 1923  \\
Average Degree (Drugs) & 2.72  \\
Average Degree (Targets) & 1.27  \\ 
Average Length of Drug SMILES & 59.06 \\
Average Length of Protein Sequence & 573.77 \\
\botrule
\end{tabular*}
\end{table}

The SMILES representation of the drugs is retrieved from the DrugBank database \cite{wishart2006drugbank}, and the contact maps of the targets are constructed from their protein sequences obtained from the UniProt database \cite{bairoch2005universal}. Our DTI network is constructed using the binary drug-target interaction matrix from the study by Yunan Luo et al. \cite{luo2017network}. 
Using the binary interaction matrix, a bipartite graph is constructed to encode the interactions between drugs and targets. We refer to this bipartite graph as the Drug-Target Interaction Graph and its formal definition is given below in Definition \ref{DTI_graph}.

\begin{figure}[!t] 
    \centering
    \includegraphics[width=1\columnwidth, keepaspectratio]{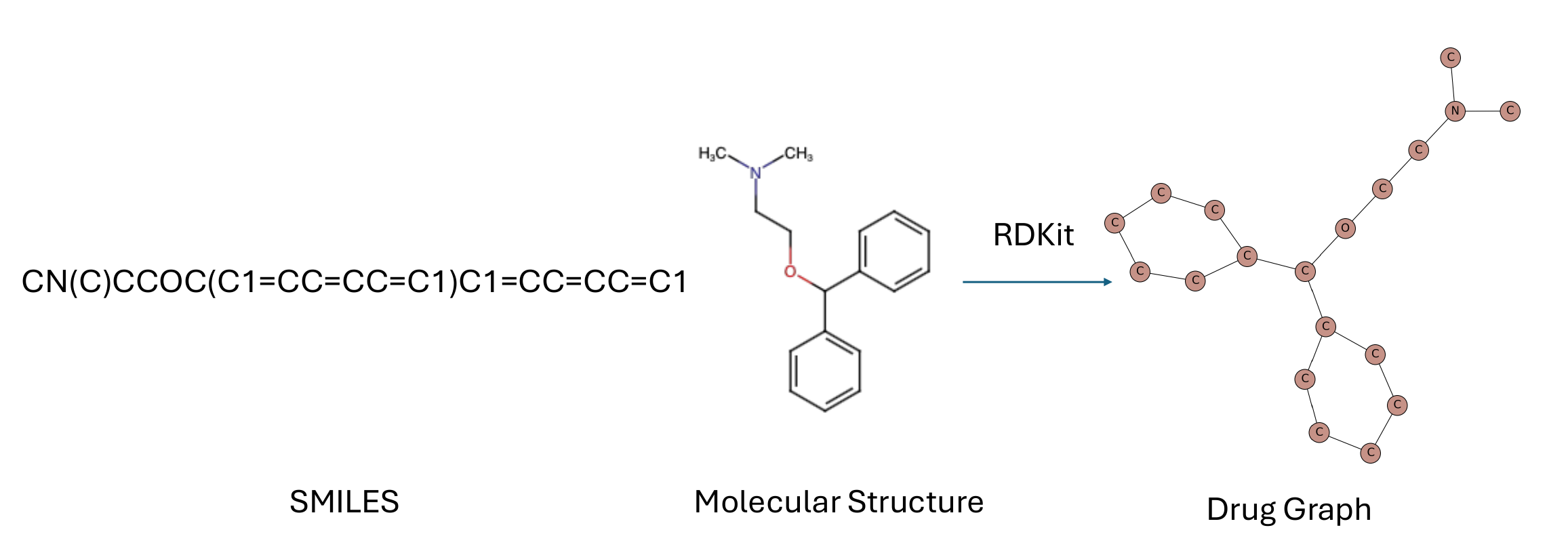}
    \caption{From SMILES expression to Drug Graph. Tools such as RDKit can be used to convert a SMILES expression to the corresponding Drug Graph. In the graph, each node represents a atom and each edge indicates the chemical bond between two atoms encoded in the SMILES expression.}
    \label{fig:drug2graph}
\end{figure}

\begin{definition}[Drug-Target Interaction (DTI) Graph]
A DTI graph $G_{dt}(V_d, V_t, E)$ is a bipartite graph,  where $V_d$ denotes the set of drugs, $V_t$ denotes the set of targets, and $E \subset V_d \times V_t$ is a set of edges representing the interactions between drugs and targets.
\label{DTI_graph}
\end{definition}

In our learning model, the structures of the drugs and targets are represented as Drug Graphs and Target Graphs, which are constructed from SMILE sequences and contact maps in our dataset. These graphs are defined in Definition \ref{def:SMILES-basedGraph} and Definition \ref{def:Sequence-basedGraph}.

\begin{definition}[Drug Graph]
    The drug graph $G(V, E, h)$ of a drug represents the structure of its molecule, where $V$ is the set of atoms and $E$ is the set of direct chemical bonds as described in the molecular expressions in SMILES. The feature matrix $h = \left(h_i, 1 \leq i \leq n \right) \in R^{n \times d}$ encodes additional chemical properties of the atoms with $h_i$ being the feature vector of the atom $v_i$.
    The drug graph is also known as the SMILES graph.
\label{def:SMILES-basedGraph}
\end{definition}

\begin{definition}[Target Graph]
    The target graph $G(V, E, h)$ of a target represents its spacial structure, where $V$ is a set of residues and $E$ is a set of edges obtained form its contact map -- a pair of nodes $\{u,v\}$ is an edge if their distance in the contact map is less than a certain threshold. The feature matrix $h = \left( h_i, 1 \leq i \leq n \right) \in R^{n \times t} $ encodes additional physicochemical properties of the residues with $h_i$ describing features of $v_i$ such as the polarity, the electrochemical properties, and the aromaticity. 
\label{def:Sequence-basedGraph}
\end{definition}

\begin{figure}[!t] 
    \centering
    \includegraphics[width=1\columnwidth, keepaspectratio]{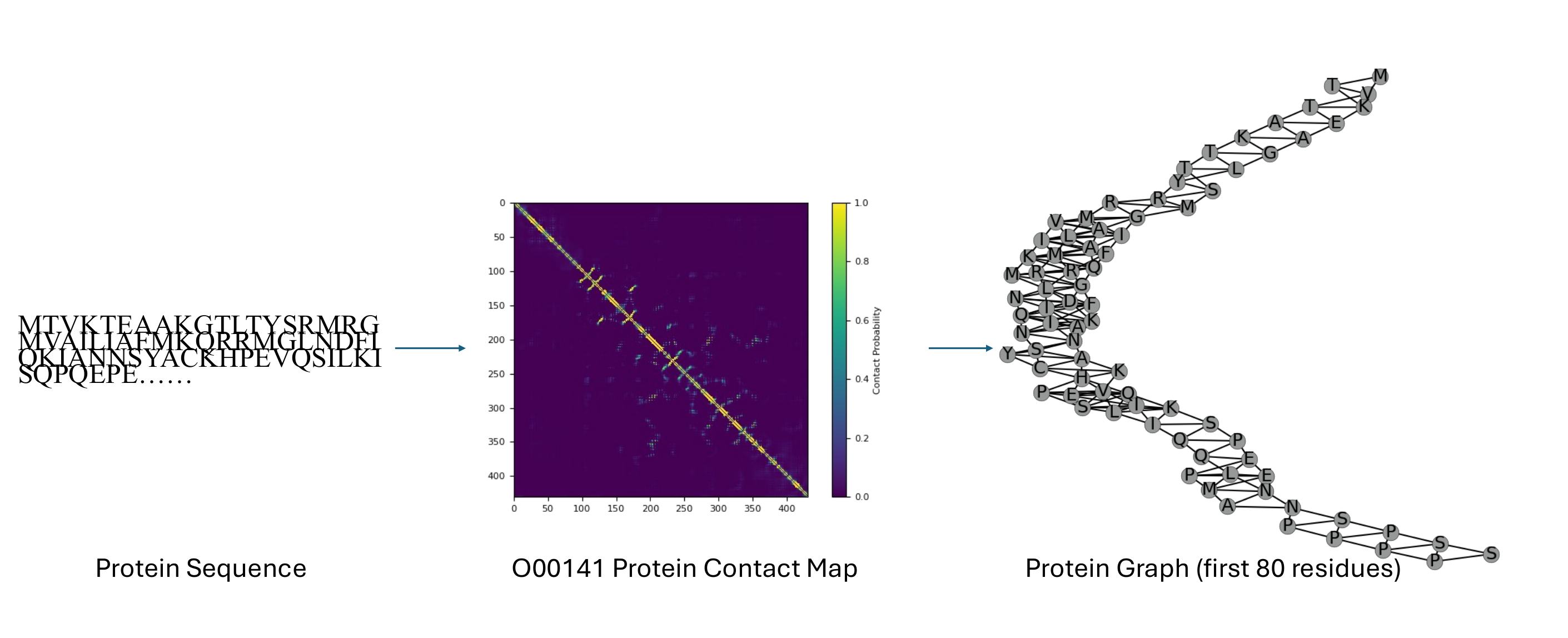}
    \caption{The protein sequence, contact map, and Target Graph of the protein O00141—Serine/Threonine‐Protein Kinase. The contact map is a matrix of contact probabilities of pairs of residues predicted by Pconsc4  \cite{michel2019pconsc4}.
    In the Target Graph, each node represents a residue and an edge between two nodes indicates that the contact probability between the two residues is above a chosen threshold.}
    \label{fig:seq2graph}
\end{figure}

Examples of Drug Graphs and Target Graphs are shown in Fig. \ref{fig:drug2graph} and Fig. \ref{fig:seq2graph}. The Fig. \ref{fig:drug2graph} shows the process of constructing the Drug Graph of a SMILES sequence. The Fig. \ref{fig:seq2graph} illustrates how the Target Graph is constructed from the protein contact map. In the figure, only the subgraph on the first 80 residues is shown.

\subsection{The Graph-in-Graph Framework for DTI Prediction}
\subsubsection{Motivations}

In the literature, distinct learning models have been used in the study of drug-target relationships, including  DTA and DTI prediction. The DTA prediction is typically modeled as a supervised learning tasks where pairs of drugs and targets with a known interaction score are training examples. Models learned in this way can be used to 
predict the affinity of completely new drug and target  pairs. Critical to this approach is the ability to represent drugs and targets numerically. While there are different ways to represent the drugs and targets,  it is a recent effort to use the embedding vectors calculated by some deep learning models on the Drug Graphs and Target Graphs and to use them in the supervised learning models \cite{nguyen2021graphdta, jiang2020drug, nguyen2021gefa, qi2024drug, jiang2022sequence}. 
 

In contrast, approaches to DTI prediction have been based on the idea of network homophily \cite{zhang2015comparison, mcpherson2001birds, gavin2002functional} in network science, suggesting that ``similar" drugs in a drug-target network tend to interact with ``similar" targets. As a consequence, learning models for DTI prediction are typically defined on a single drug-target network, emphasizing the use of known interactions within the network to predict new drug-target interactions in the same network. From the perspective of machine learning, these models for DTI prediction are transductive learning models and are trained in an unsupervised or semi-supervised setting. Much effort has been put into improving the effectiveness of this approach by incorporating additional information to measure the similarity between drug pairs and target pairs. Good examples of such techniques are that construct additional  network layers of drugs and targets and use GNNs and matrix decomposition techniques to 
handle the resulting heterogeneous network \cite{zheng2013collaborative, luo2017network, liu2021gadti, wang2014drug, wan2019neodti}.   
We note, however, that no methods have been proposed that can make use of the structure of the drugs and targets directly in a learning model.


In this paper, we propose a framework in which one can develop models to simultaneously learn from detailed molecular features (as in DTA) and global interaction networks (as in DTI). Models based on this framework provide a better approach to the integration of interaction and molecular data for more effective DTI prediction. We discuss our framework and model in the rest of this section.

\subsubsection{A Hierarchical GNN for DTI}
A key challenge arises from the heterogeneous nature of drug and target representations. Drug Graphs are based on chemical structure derived from the SMILES strings, while Target Graphs are constructed from protein sequence-based features and contact maps. Integrating such distinct graph types into a cohesive predictive model necessitates a mechanism capable of unifying these diverse representations.

\begin{definition}
A GiG model \( GiG(\mathcal{D}, \mathcal{T}, E) \) is a 2-layer hierarchical GNN on a set of \( m \) drugs and \( n \) targets. Here, \( \mathcal{D}=\{D_i, 1 \leq i \leq m\} \) is a set of Drug Graphs for drugs with \( D_i=G(V_{D_i}, E_{D_i}) \) 
being the Drug Graph of drug $i$ and \( \mathcal{T}=\{T_i, 1 \leq i \leq n\} \) represents a collection of Target Graphs for targets with
\( T_i=G(V_{T_i}, E_{T_i}) \) the Target Graph of target $i$.
The set of edges \( E \subseteq \mathcal{D} \times \mathcal{T}\) denote potential interactions between drugs and targets. When these Drug Graphs and Target Graphs are treated as meta nodes, \( GiG(\mathcal{D}, \mathcal{T}, E) \) has a bipartite graph structure.
\label{gig}
\end{definition}

\begin{figure*}[!t] 
    \centering
    \includegraphics[width=0.85\textwidth, keepaspectratio]{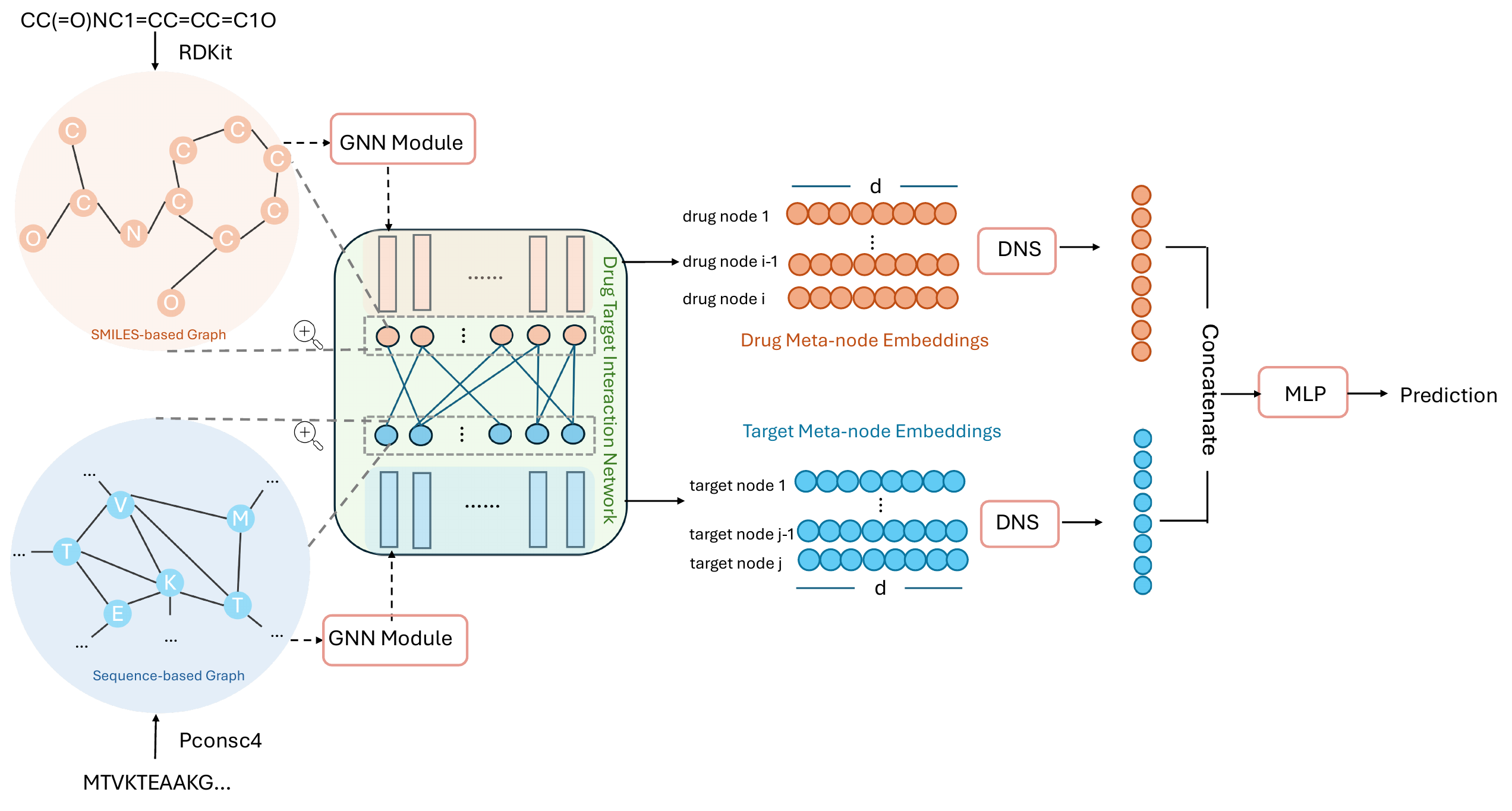}
    \caption{Overview of the GiG model. The left panel illustrates Drug Graphs for drugs derived from SMILES representations, while the right panel shows Target Graphs for targets. The square panel represents the drug-target interaction (DTI) graph, modeled as a bipartite network where drug and target meta-nodes interact. Features extracted from Drug and Target graphs are integrated into the main graph, enabling a hierarchical learning framework that captures both local molecular structures and global interaction patterns for improved DTI prediction.} 
    \label{fig:gig}
\end{figure*}

As illustrated in Figure \ref{fig:gig}, the GiG model incorporates three intertwine graph structures: the drug-target network $GiG(\mathcal{D}, \mathcal{T}, E)$, the Drug Graphs $\{D_i=G(V_{D_i}, E_{D_i}), 1 \leq i \leq m\} $, and the Target Graphs $\{T_i=G(V_{T_i}, E_{T_i}), 1 \leq i \leq n\}$. The core component, $GiG(\mathcal{D}, \mathcal{T}, E)$, is a bipartite graph where drugs and targets are represented as nodes and their known or potential interactions form the edges.

This hierarchical design empowers the GiG model to learn from both fine-grained structural information specific to individual drugs and targets and broader interaction-level patterns across the network. Unlike traditional DTI models that rely on precomputed similarity measures, the GiG model learns directly from the raw molecular and sequence data to generate task-optimized embeddings. This enables the model to capture interaction-relevant features, such as recurring substructures in drug molecules and conserved motifs within protein sequences, which are often critical for binding specificity.

In the GiG model, there are three GNN modules: GNN on Drug Graphs, GNN on Target Graphs and GNN on the DTI network. The embeddings learned by Drug GNN and Target GNN capture the features of biochemical structures of the drugs and targets. These embeddings then used to initialize the meta-node attributes in the Drug-Target GNN on the bipartite DTI graph. This fusion of micro-scale (molecular/sequence) and macro-scale (interaction) features enhances the predictive accuracy by learning the representations from both the biological structure and the network topology. 

\subsubsection{Message Passing in GiG Model}
The GiG employs an iterative message-passing mechanism that alternates between the updates of the Drug GNN, the Target GNN and the Drug-Target GNN. In each iteration, node features of the Drug Graphs and the Target Graphs are transformed by their respective GNNs and aggregated via a global pooling function to produce graph embeddings. These embeddings are used to initialize the features of meta-nodes in the DTI graph and the Drug-Target GNN propagates information across the bipartite structure, refining the meta‐node embeddings to capture cross‐domain interactions. These updated meta‐node embeddings are passed into the prediction layers to compute interaction scores and the corresponding loss. During backpropagation, the gradients flow from the loss through the Drug–Target GNN and into the Drug and Target GNNs to enhance the representations at the node level, creating a bidirectional feedback loop that progressively integrates local biochemical information and global network information. 

In the following, we formally define the message passage mechanisms for these three GNNs. Let \(\mathbf{X}_{D}^{0}\) and \(\mathbf{X}_{T}^{0}\) be the initial node feature matrices for a Drug Graph $D$ and a Target Graph $T$, respectively. Each column of $\mathbf{X}_{D}^0 = \left(x_{D,1}, x_{D,2}, ...,x_{D,m} \right) $ is the feature vector of an atom comprising one-hot atom symbols, atom degree, total hydrogens, implicit valence, and an aromaticity flag. A column of $\mathbf{X}_{T}^0 = \left(x_{T,1}, x_{T,2}, ...,x_{T,n} \right)$ encodes the features of a residue, using five binary class indicators (aliphatic, aromatic, polar neutral, acidic charged, basic charged) and seven physicochemical descriptors (molecular weight, pKa, pKb, pKx, isoelectric point, hydrophobicity at pH2, hydrophobicity at pH7).

\begin{itemize}
\item The Message Passing Mechanism of the Drug GNN.

The Drug GNN implements message passing by updating the embedding of each atom by aggregating its own features with those of its bonded neighbors, thus encoding the bonding context of each atom within the molecular graph
    \begin{equation}
        h_{D,u}^{k} = \sigma \left( W_{D}^{k} \sum_{v \in N(u) \cup \{u\}} h_{D,v}^{k-1} + b^{k} \right)
    \label{eq:drug_gnn_mp}
    \end{equation}
Here, $W_D^{k}$ is the learnable weight matrix applied to the normalized sum of the self and neighbor embeddings at $k-1$ layer, and $b^{k}$ is a bias vector, and $\sigma(\cdot)$ denotes the ReLU activation function. This update is applied iteratively for the three layers, which expands the receptive field of the neuron for each atom to encode longer-range chemical interactions. In this way, our Drug GNN model can encode relevant functional groups within each molecule.

\item The Message  Passing  Mechanism of the Target GNN

The Target GNN implements message passing by updating each residue’s embedding through aggregating its own features with those of spatially adjacent residues (as defined by the contact map), thereby encoding each residue’s structural neighborhood. 
    \begin{equation}
        h_{T,u}^{k} = \sigma \left( W_{T}^{k} \sum_{v \in N(u) \cup \{u\}} h_{T,v}^{k-1} + b^{k} \right)
    \label{eq:target_gnn_mp}
    \end{equation}
All the parameters are as defined in Eq. \ref{eq:drug_gnn_mp}. Iterating this update over three layers allows each residue to aggregate information from nodes up to 3-hop away. In practice, this means that a given node not only “sees” its immediate contacts, but also learns about long-range structural motifs in its embedding.

\item The Message Passing Mechanism of Drug-Target GNN

Let $\mathbf{H}_{DT}^{k}= [\mathbf{H}_{\mathcal{D}}^{k}, \mathbf{H}_{\mathcal{T}}^{k}]^T$ be the feature matrix of the Drug-Target GNN in layer $k$, where   
\[
\begin{aligned}
    \mathbf{H}_{\mathcal{D}}^k = \begin{bmatrix} h_{D_1}^{k}  \\ h_{D_2}^{k} \\ \vdots \\ h_{D_m}^{k} \end{bmatrix}
    \quad \textnormal{and} \quad
    \mathbf{H}_{\mathcal{T}}^k = \begin{bmatrix} h_{T_1}^{k}  \\ h_{T_2}^{k} \\ \vdots \\ h_{T_n}^{k} \end{bmatrix}
\end{aligned}
\]
are respectively the features of the drug meta nodes and  the target meta nodes. In $\mathbf{H}_{\mathcal{D}}^k$ and $\mathbf{H}_{\mathcal{T}}^k$, $h_{D_i}^{k}$ is the feature vector of the drug meta node $D_i$ and $h_{T_i}^{k}$ is the feature vector of the target meta node $T_i$.

The Drug–Target GNN runs the message passing on the bipartite interaction graph $G_{dt} \left( \mathcal{D}, \mathcal{T}, E \right)$: each drug (or target) node updates its embedding by aggregating its own features with its neighbours in the other partite set of the graph $G_{dt} \left( \mathcal{D}, \mathcal{T}, E \right)$ and thereby encodes the cross-domain interaction context. In matrix form, the update equations of message passing mechanism of the Drug-Target GNN are
\begin{equation}
    \mathbf{H}_\mathcal{D}^{k} = \mathcal{MP}_{main}\left( \mathbf{H}_\mathcal{D}^{k-1}, \mathbf{W}_{self}^{k}; \; 
    \mathbf{H}_\mathcal{T}^{k-1}, \mathbf{W}_{neigh}^{k} \right) 
    \label{eq:drug_update}
\end{equation}
and
\begin{equation}
    \mathbf{H}_\mathcal{T}^{k} = \mathcal{MP}_{main}\left( \mathbf{H}_\mathcal{T}^{k-1}, \mathbf{W}_{self}^{k}; \;
    \mathbf{H}_\mathcal{D}^{k-1}, \mathbf{W}_{neigh}^{k} \right).
    \label{eq:target_update}
\end{equation}
The update equation of the feature of a general node $u$ in the bipartite graph is
    \begin{equation}
        h_{DT,u}^{k} = \sigma (\alpha_{u,u} W_{self}^{k}h_{DT,u}^{k-1}+ \sum_{v \in N(u)} \alpha_{u,v} W_{neigh}^{k} h_{DT,v}^{k-1} + b^{k})
    \end{equation}
Here, $\alpha_{u,u}$ denotes the attention coefficient of node $u$ with itself, and $\alpha_{u,v}$ denotes the attention coefficient from neighbor $v$ to $u$; both are computed via a shared trainable attention mechanism and normalized together by the softmax function. The weight matrices for layer $k$, $W_{\text{self}}^{k}$ and $W_{\text{neigh}}^{k}$, applied to the “self” node and its neighbors, respectively. 
\end{itemize}



One of the unique features of our model is that the three GNNs are intertwined through the interactions of the message passing mechanisms:
\begin{itemize}
\item The feature matrix of the Drug-Target GNN is initialized by using the global mean pooling function $gep$ to produce the graph level embeddings from the node embeddings of the Drug GNN and the Target GNN, i.e. 
\[
{h}_{D_i}^{0} = \mathbf{gep}\left( \mathcal{G}_{D_i} \right) = mean( h_{D,u},u \in V_D)
\] and 
\[
{h}_{T_i}^{0} = \mathbf{gep}\left( \mathcal{G}_{T_i} \right) = mean( h_{T,u},u \in V_T).
\]This type of data information fusion is supported by the message passing mechanism designed in the GiG model.

\item The same loss function is used by the three GNNs in the backpropagation during training. Our experiments show that even if this loss function does not measure directly the quality of the node embeddings, the Drug GNN and Target GNN produce embeddings that are better than the node embedding techniques used in baseline models such as GCN, GAT, and Node2Vec.
\end{itemize}


Our model differs from previous heterogeneous GNNs in that its Drug–Target GNN is built on a bipartite graph and directly incorporates drug–drug and target–target similarities through message passing. By using at least two layers, each drug or target meta-node collects information from both its one-hop and two-hop neighbors. Because two-hop neighbors share the same type as the meta-node itself, the final embeddings capture both cross-domain interactions and within-domain similarities.


\subsubsection{Predicting the Interactions}
The final layer of the Drug-Target GNN is a multilayer perceptron (MLP) with a sigmoid output function that takes as input the concatenated embeddings of each drug–target pair and outputs a scalar interaction score.  We use the MLP to predict DTIs. For each pair of drug and target meta-nodes, we concatenate their embeddings and feed it into the MLP:
\begin{equation}
\hat{y}_{ij} = \sigma_{sigmoid} \left( \mathbf{W}_2 \cdot \sigma(\mathbf{W}_1 [\mathbf{Z}_{D_i}^k \, \| \, \mathbf{Z}_{T_j}^k] + \mathbf{b}_1 ) + \mathbf{b}_2 \right)
\end{equation}
Here, \([\cdot \, \| \, \cdot]\) denotes the concatenation of drug and target embeddings,  \(\mathbf{W}_1\) and \(\mathbf{W}_2\) are learnable weight matrices,  \(\mathbf{b}_1\) and \(\mathbf{b}_2\) are biases, $\sigma$ is the ReLU activation function $ReLU(x)=max(0,x)$.

The GiG model is trained by using binary cross-entropy loss defined as
\begin{equation}
\mathcal{L} = -\frac{1}{N} \sum_{(i,j)} \left[ y_{ij} \cdot \log(\hat{y}_{ij}) + (1 - y_{ij}) \cdot \log(1 - \hat{y}_{ij}) \right]
\label{eq:loss}
\end{equation}
Here ${y}_{ij}=1$ indicate a positive interaction between drug $i$ and target $j$ and ${y}_{ij}=0$ indicate the pair $(i,j)$ is a hard negative sample. The value $\hat{y}_{ij}$ is the predicted interaction score.

The value of the loss function depends not only on the topological patterns of the drug-target network, but also on the molecular features of individual drugs and targets. Therefore, the back-propagation process during the training of our model is able to refine the underlying representation of both the DTI graph and the molecular graphs. This recursive refinement mechanism ensures that the final embeddings integrate both the molecular-scale details and the network-level interaction context, leading to robust and generalizable predictions.

We use hard negative sampling \cite{zhang2013optimizing, yang2024does} during MLP training. After the MLP has predicted the scores for each drug–target pair, we sort these negative pairs by their predicted scores in descending order and then select the top-ranked pairs, those that the model is most likely to misclassify as positive, as the hard negative samples. The total number of hard negatives is set according to a fixed ratio (1:1) relative to the number of positive samples. By forcing the model to learn from these “difficult” false positives, we sharpen its ability to distinguish true interactions from non‐interactions.
To allow the model to learn an adaptive decision threshold, we replace a standard sigmoid with a shifted sigmoid function \[\sigma_{\text{shift}}(x) = \frac{1}{1 + \exp\left(-\frac{a x + b}{t}\right)}.\]
The three parameters \(a\), \(b\) and \(t\) are optimized in the training process, allowing the GiG model to adaptively learn optimal decision boundaries that are better suited for DTI predictions.

The GiG model is designed to overcome the limitations identified in baseline methods (Section ``The Baseline") and unsupervised node embedding approaches (Section ``Node2Vec for Feature Extraction"). The baseline models that rely solely on graph connectivity lack the biochemical and structural information embedded in the molecular graphs of drugs and targets. The unsupervised learning approach based on node2vec requires a pre-training step to generate node embeddings, and its learned representations are separated from the end-to-end learning process, limiting the integration of local structural information with global network interactions.

In contrast, GiG is a fully end-to-end model that directly integrates the Drug Graphs and the Target Graphs into a unified learning framework. This is achieved by constructing two levels of graphs: molecular graphs representing individual drugs and targets, and a main graph representing the drug-target interactions. Each drug is represented as a Drug Graph based on its SMILES sequence, capturing atomic interactions and bond structures. Similarly, each target is represented as a Target Graph, which encodes spatial relationships derived from its contact maps. These molecular graphs are processed using GNNs to extract meaningful representations that capture local structural and chemical features.

Our GiG processes the graph features in real time, allowing for joint optimization of node embeddings and interaction predictions. After processing the molecular graphs by the Drug GNN and Target GNN, the learned embeddings are treated as meta-nodes features in the Drug-Target GNN, where the DTI network is modeled as a bipartite graph. Through hierarchical message passing, GiG enables multi-scale feature propagation: molecular GNNs capture local dependencies, while the main graph GNN models global interaction patterns. This dual-level learning mechanism enhances the model’s capacity to identify subtle interaction cues that baseline methods and unsupervised node embeddings often miss.

Moreover, the end-to-end nature of GiG allows it to optimize molecular and interaction-based embeddings simultaneously, leading to more robust and interpretable predictions. The 2-leavel hierarchical structure in the GiG makes it possible to seamlessly take advantages of the strength of both transductive and inductive learning. On the one hand, the learning of the Drug-Target GNN captures existing interaction patterns, while on the other hand the collection of molecular graphs are used to train a common Drug GNN and Target GNN inductively.

In the following sections, we describe the experimental setup, implementation details, and baseline comparisons, followed by a comprehensive analysis of the model's predictive performance and its contributions to DTI prediction.

\section{Experimental Setup} 
In this section, we present the experimental setup used to evaluate our GiG framework for DTI prediction. Our experiments are designed to systematically compare different models, including baseline GNN models, node2vec-based feature extraction, and the proposed GiG framework.

In our experiments, we used several models as baselines, which serve as reference points for evaluating the effectiveness of hierarchical graph modeling. To explore the benefits of unsupervised feature learning, we apply node2vec to extract embeddings from Drug and Target Graphs, providing an alternative to supervised GNN-based feature extraction. We evaluated the full GiG model, which integrates detailed molecular and sequence-level information with the global interaction graph through a hierarchical structure.

We use the same dataset and hyperparameters to train all models to ensure a fair and objective comparison. The experimental results offer insights into the advantages of combining molecular graphs with interaction networks and highlight the effectiveness of the GiG framework for DTI prediction.

\subsection{The Baseline}
\label{Section 3.1}
For the baseline comparison, we designed a simple GNN-based model that utilizes randomly initialized node features instead of structural or sequence-based information. Specifically, each drug and target node in the bipartite interaction graph is assigned a feature vector sampled from a standard normal distribution.

The randomly generated node features are then processed through either a GCN or a GAT model to learn node embeddings based on the connectivity structure of the bipartite interaction graph. After obtaining the node embeddings, drug--target pairs are constructed according to the known interaction edges. For each drug--target pair, their corresponding embeddings are concatenated and passed through a MLP to predict the probability of interaction.

This baseline setting provides a measure of how much predictive performance can be achieved purely from the graph topology without incorporating explicit molecular or sequence-level features. It also establishes a reference point for evaluating the contributions of hierarchical graph representations within the GiG framework.

\subsection{Node2Vec for Feature Extraction}
\label{Section 3.2}
To explore the effectiveness of unsupervised feature extraction for molecular graphs, we employ \textit{node2vec} model \cite{grover2016node2vec} to generate node embeddings for both Drug and Target Graphs. The Node2vec is a random-walk-based embedding technique that captures neighborhood information through biased random walks, optimizing for both homophily and structural equivalence.

The Node2vec performs random walks of length \(l\) from each node of a  graph and generates context sequences that resemble sentences in natural language processing. In our experiments, we set the walk length \(l\) to 20, the number of walks per node to 10, and the context size (window size) \(w\) to 10, ensuring sufficient exploration of local neighborhoods. Following standard practice for efficient contrastive learning, the number of negative samples per node is set to 1.

The Node2vec sampling strategy is controlled by two hyperparameters: \(p\) and \(q\). Both are set to 1.0 in our configuration, corresponding to unbiased random walks that equally favor breadth-first and depth-first search strategies. This choice allows the model to capture both local structures and broader connectivity patterns in the molecular graphs.

The optimization objective of Node2vec maximizes the likelihood of observing a node's neighbors within a specified window size \(w\):

\[
\max \prod_{u \in V} \prod_{v \in \mathcal{N}_w(u)} \Pr(v \mid u),
\]
where \(\mathcal{N}_w(u)\) represents the neighborhood of node \(u\) within window size \(w\), and \(\Pr(v \mid u)\) is the conditional probability of visiting node \(v\) given node \(u\).

The node embeddings learned through Node2vec, denoted as \(\mathbf{Z}_D\) and \(\mathbf{Z}_T\), are directly utilized as feature representations for the drug and target meta-nodes in the bipartite interaction graph. These embeddings are then passed to a GNN-based model for DTI network. Unlike the purely random initialization used in the baseline, the use of Node2vec embedding enhance the model's capacity to distinguish meaningful structural patterns by capturing topological dependencies.

These baseline models provide a benchmark for evaluating the benefit of using molecular structures in the GiG framework over unsupervised structure representation.

\subsection{The Graph-in-Graph Model}
\label{sec:GiG}
To address the shortcomings of the baselines (Section ``The Baseline") and unsupervised embedding approaches (Section ``Node2vec for Feature Extraction"), we propose the GiG model. Traditional graph methods that rely solely on topology overlook the rich biochemical and structural cues embedded in Drug Graphs and Target Graphs, while Node2vec-based pipelines require a separate pretraining phase and yield embeddings that are decoupled from the downstream learning objective, limiting the integration of local molecular graph features with global DTI patterns.

In contrast, GiG is a fully end-to-end model that directly integrates Drug Graphs and Target Graphs into a unified learning framework. It is implemented in Python 3.9.0 using PyTorch 2.2.0 (CUDA 11.8, cuDNN 8.7.0) and PyTorch Geometric 2.5.3. All experiments are conducted on a single NVIDIA Tesla V100-SXM2-32GB GPU with driver version 550.163.01 and CUDA 12.4.


Training is conducted with a batch size of 64 for both Drug and Target GNNs, using Adam optimization (learning rate \(1 \times 10^{-4}\)) for up to 1000 epochs. The dropout (\(p = 0.2\)) is applied after every GNN layer in the Drug GNN, Target GNN, and Drug-Target GNN to mitigate overfitting. A hard-negative sampling strategy ensures a 1:1 balance of positive and negative examples within each batch.

\subsection{Performance Evaluation Metrics}

 To evaluate the performance of the models, we selected four key metrics: the area under the receiver operating characteristic curve (AUC) \cite{provost1998case}, the area under the precision-recall curve (AUPRC) \cite{raghavan1989critical, sokolova2009systematic}, the F1 Score \cite{lipton2014thresholding}, and the Matthews Correlation Coefficient (MCC) \cite{chicco2020advantages}. These metrics were chosen to provide a comprehensive view of the model's predictive capabilities, especially considering the use of hard negative sampling during the training phase.

\textbf{Area Under the ROC Curve (AUC)}

The AUC measures the model's ability to distinguish between positive (interacting drug-target pairs) and negative (non-interacting drug-target pairs) samples across various classification thresholds. It is calculated by plotting the True Positive Rate (TPR) against the False Positive Rate (FPR) and computing the area under the resulting curve. A higher AUC score reflects the model's ability to rank positive instances higher than negative ones consistently, indicating effective global separability. Importantly, the AUC remains robust even with hard negative sampling, as it evaluates relative ranking rather than absolute prediction scores.

\textbf{Area Under the Precision-Recall Curve (AUPRC)}

The AUPRC evaluates the relationship between precision (the ratio of true positives to all predicted positives) and recall (sensitivity) across different thresholds. This metric is particularly meaningful in DTI prediction due to the sparsity of true interactions compared to the potential negative pairs. By focusing on the model's performance with respect to the positive class, the AUPRC provides insights into how well the model avoids false positives while capturing true interactions. Given the use of hard negative sampling, the AUPRC helps measure the model's precision in distinguishing true positives from challenging negatives.

\textbf{F1 Score}

The F1 score is the harmonic mean of Precision and Recall, reflecting the model's ability to maintain a balance between capturing true positives and minimizing false positives. In our experimental setup, the MLP classifier is trained with a 1:1 ratio of hard negatives to positives, making the F1 Score a meaningful indicator of the classifier's discriminative power in a balanced sampling context.

\textbf{Matthews Correlation Coefficient (MCC)}

The MCC offers a more nuanced view of the model's predictive consistency across both positive and hard negative samples. A higher MCC value indicates stronger overall performance in correctly identifying interactions and non-interactions.

By employing these four evaluation metrics, we ensured a comprehensive evaluation of the GiG model's performance. The AUC and the AUPRC provide insights into the global and class-specific ranking capabilities, while the F1 Score and the MCC reflect the classifier's decision boundary quality under hard negative sampling. Together, these metrics present a holistic view of the strength of the model in the DTI prediction.

\section{Results and Discussion}
In this section, we present the experimental results on the GiG model and compare its performance with baseline methods under various settings. We evaluated the model's robustness across different GNN configurations and benchmark it against traditional GNN-based DTI prediction models.

\subsection{Comparison with Baseline Methods}
\label{section:A}
To demonstrate the effectiveness of the GiG model, we compare it with two baseline methods introduced in the experimental setup: DTI-GCN, DTI-GAT, and the Node2vec-enhanced GNNs. These models are traditional graph-based DTI prediction approaches, leveraging standard GCN and GAT architectures with either randomly initialized node features or unsupervised embeddings derived from Node2vec, and we used them as benchmarks in the comparison. While GCN and GAT directly process raw graph structures to learn node representations, Node2vec enhances the representation power by capturing higher-order proximity through random walks, improving the expressiveness of node embeddings in downstream tasks.

All the models in this comparison employed the shifted sigmoid activation function in their final prediction layer. 
The four evaluation metrics that we used, AUC, AUPRC, F1 Score, and MCC provide a comprehensive view of each model's predictive capability. The results under the 7:1:2 train-validation-test split are presented in Table \ref{tab:method_performance}.

\begin{table*}[t]
\caption{Performance comparison of various models in the drug-target interaction prediction task.\label{tab:method_performance}}
\tabcolsep=0pt
\begin{tabular*}{\textwidth}{@{\extracolsep{\fill}}lcccc@{\extracolsep{\fill}}}
\toprule%
\textbf{Method} & \textbf{ROC-AUC} & \textbf{AUC-PR} & \textbf{F1 Score} & \textbf{MCC} \\
\midrule
DTI-GCN \cite{kipf2016semi}  & 0.9032 & 0.8318 & 0.4305 & 0.3300 \\
DTI-GAT \cite{velivckovic2017graph} & 0.9652 & 0.9633 & 0.7000 & 0.5935 \\
\midrule
Node2Vec-5-enhanced GCN \cite{grover2016node2vec}  & 0.9391 & 0.9147 & 0.8594 & 0.7223 \\
Node2Vec-10-enhanced GCN \cite{grover2016node2vec} & 0.9582 & 0.9241 & 0.9044 & 0.8078 \\
Node2Vec-5-enhanced GAT \cite{grover2016node2vec} & 0.9534 & 0.8935 & 0.9394 & 0.8772 \\
Node2Vec-10-enhanced GAT \cite{grover2016node2vec} & 0.9595 & 0.8992 & 0.9564 & 0.9124 \\
\midrule
GIG [GCN-GCN][GAT]  & \textbf{0.9934} & \textbf{0.9910} & \textbf{0.9923} & \textbf{0.9845} \\
\botrule
\end{tabular*}

\begin{tablenotes}%

\item Note: The Model performance on the DTI prediction task using a 7:1:2 train-validation-test split. “Node2Vec-5” and “Node2Vec-10” denote Node2Vec embeddings trained for 5 and 10 epochs, respectively. The proposed GiG model with the [GCN-GCN][GAT] configuration achieves the highest performance across all metrics, demonstrating its ability to capture multi-scale interaction features. Here, [GCN-GCN][GAT] refers to two parallel GCNs for extracting Drug and Target Graph embeddings, followed by a GAT for feature learning on the DTI network.
\end{tablenotes}
\end{table*}

When compared to the baselines, the [GCN-GCN][GAT] configuration of GiG achieves superior performance across all evaluation metrics, recording an AUC of 0.9934, an AUPRC of 0.9910, an F1 Score of 0.9923, and an MCC of 0.9845. Here, [GCN-GCN][GAT] refers to a configuration where two parallel GCNs are used to extract features independently from Drug and Target Graphs, while a GAT is employed to model the DTI graph. These results highlight the strength of GiG's hierarchical message-passing mechanism, which enables it to capture not only local structural dependencies within molecular graphs but also long-range interaction patterns in the main DTI network. This multi-scale aggregation allows for enriched feature propagation, enhancing the model's discriminative power for link prediction.

Moreover, the AUC and AUPRC scores achieved by GiG provide further evidence of its superior ranking capability. Unlike threshold-dependent metrics, both the AUC and the AUPRC evaluate the model's discriminative power across all possible decision thresholds. The GiG's near-perfect 0.9934 AUC and 0.9910 AUPRC demonstrate its ability to rank true interactions significantly higher than non-interactions over a wide range of cutoff values. This threshold-independence is particularly important in real-world DTI predictions, where selecting an optimal threshold is challenging due to the heterogeneity of biological networks. The ability of GiG to maintain high ranking fidelity without fine-tuning thresholds underscores its practical applicability in large-scale drug discovery tasks.

One of the important factors that contributes to the power and effectiveness of our GiG model is the incorporation of both transductive and inductive learning paradigms. The transductive learning leverages existing DTI network information to capture established interaction patterns, while the inductive learning enables the GiG model to learn a common embedding model in a supervised way for the drug meta-nodes and the targets meta-nodes from a collection of Drug Graphs and Target Graphs. This learning strategy has a significant advantage over traditional models, since it used both the molecular graphs of drugs and targets, and the global DTI graph within the same hierarchical architecture, our GiG is able to harness both local molecular structural information and global interaction patterns, significantly enhancing its predictive power.

\begin{figure*}[!t]
  \centering
  \begin{minipage}[t]{0.48\textwidth}
    \centering
    \includegraphics[width=\linewidth,height=5cm]{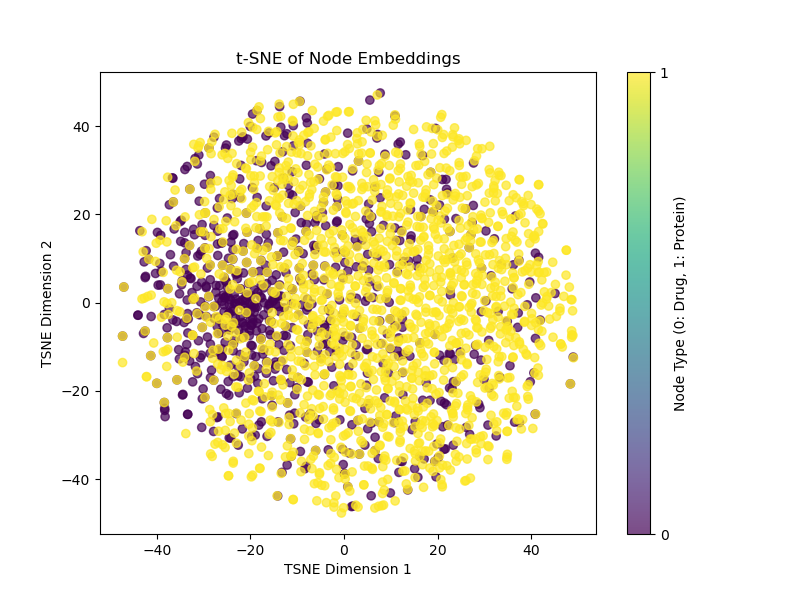}
    \\[-0.7em]\small (a) DTI-GCN
  \end{minipage}\hfill
  \begin{minipage}[t]{0.48\textwidth}
    \centering
    \includegraphics[width=\linewidth,height=5cm]{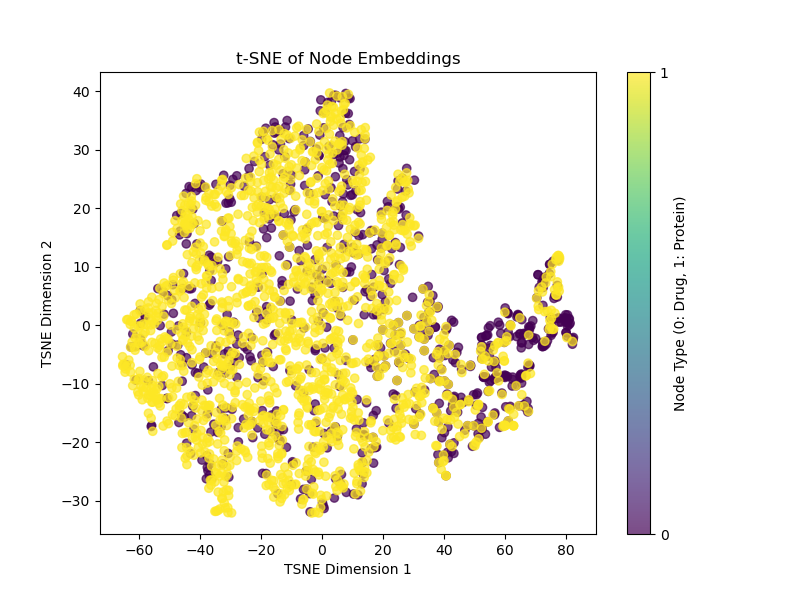}
    \\[-0.7em]\small (b) DTI-GAT
  \end{minipage}

  \vspace{0.8em}

  \begin{minipage}[t]{0.48\textwidth}
    \centering
    \includegraphics[width=\linewidth,height=5cm]{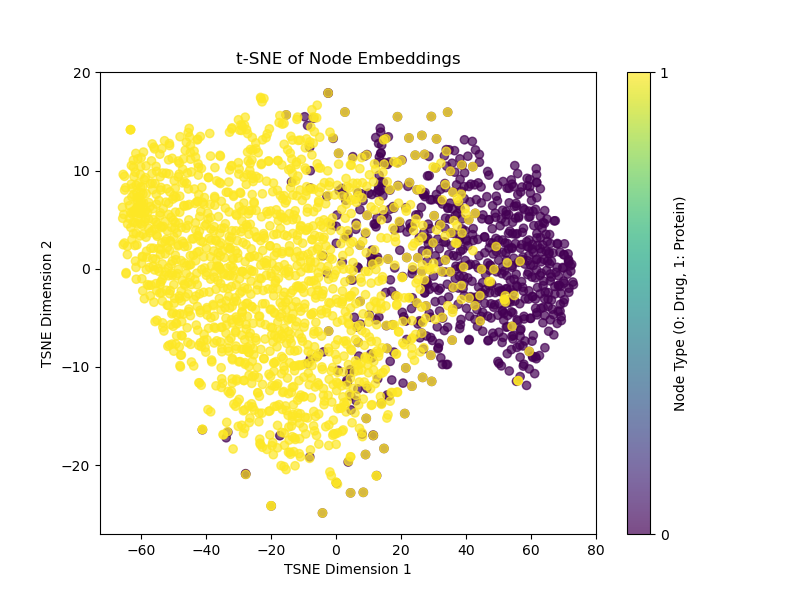}
    \\[-0.7em]\small (c) Node2Vec-GCN
  \end{minipage}\hfill
  \begin{minipage}[t]{0.48\textwidth}
    \centering
    \includegraphics[width=\linewidth,height=5cm]{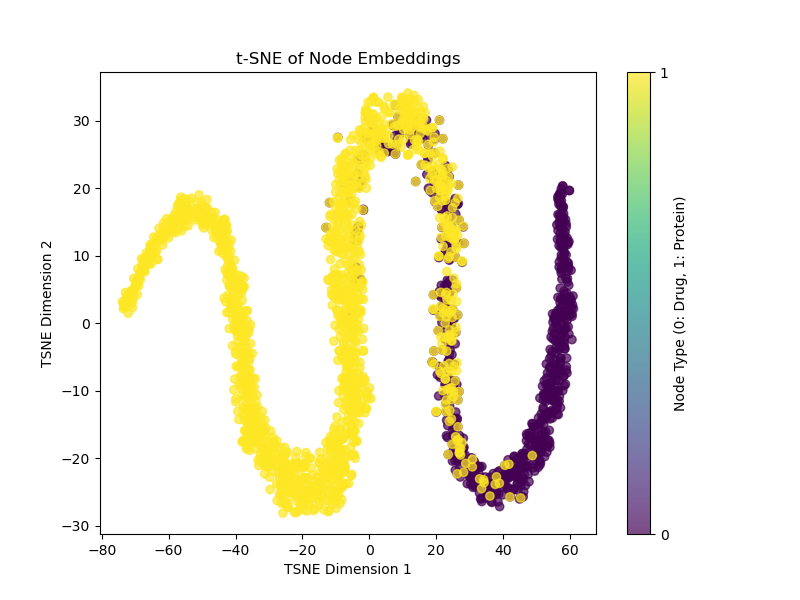}
    \\[-0.7em]\small (d) Node2Vec-GAT
  \end{minipage}

  \vspace{0.8em}

  \begin{minipage}[t]{0.48\textwidth}
    \centering
    \includegraphics[width=\linewidth,height=5cm]{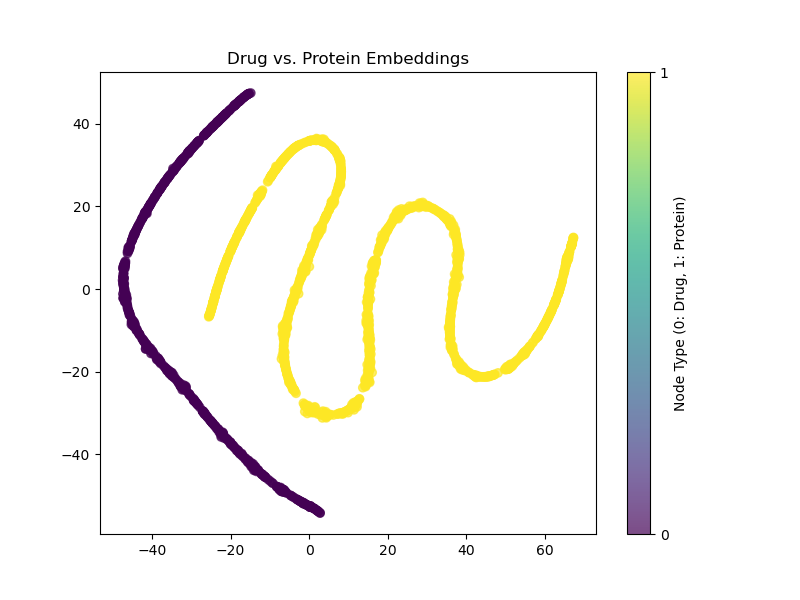}
    \\[-0.7em]\small (e) GiG Model
  \end{minipage}

  \caption{Comparative t-SNE visualizations from five models: (a) DTI-GCN, (b) DTI-GAT, (c) Node2Vec-GCN, (d) Node2Vec-GAT, and (e) GiG model. Purple nodes: drug; yellow nodes: protein.}
  \label{fig:tsne_comparison}
\end{figure*}





To further validate the representational power of each model's GNN-based feature extraction, we performed t-SNE visualization on the node embeddings of drugs and targets before they are passed into the MLP classifier. The Fig. \ref{fig:tsne_comparison} demonstrates that only the GiG model ([GCN-GCN][GAT]) can effectively separate drug and target nodes into distinct clusters, while the baseline models struggle to achieve this level of separation.

From these images, we can observe that DTI-GCN and DTI-GAT (Figures \ref{fig:tsne_comparison}(a) and \ref{fig:tsne_comparison}(b)) present a largely mixed distribution of drug and target embeddings, indicating suboptimal feature separation. The Node2Vec-enhanced GCN and GAT (Figures \ref{fig:tsne_comparison}(c) and \ref{fig:tsne_comparison}(d)) slightly improve the clustering, yet significant overlap remains. In contrast, the [GCN-GCN][GAT] configuration from GiG (Figure \ref{fig:tsne_comparison}(e)) produces a clear and well-defined bifurcation, where drug nodes (purple) and target nodes (yellow) are distinctly mapped to separate manifolds. This demonstrates GiG's capacity for capturing multi-scale topological dependencies, which are crucial for accurate DTI prediction. Furthermore, this distinct separation of drugs and targets reflects GiG's ability to implicitly capture the underlying structure of Drug-Drug Interactions (DDI) and Protein-Protein Interactions (PPI) without explicitly modeling them, contrasting with traditional approaches that rely on predefined interaction networks. Moreover, the clear separation of drugs and targets into distinct clusters suggests that GiG's hierarchical message-passing mechanism could be extended to more specialized tasks such as DDI and PPI prediction, leveraging its ability to take the advantage of molecular-level structures.

The visualization analysis strongly supports the quantitative results in Table \ref{tab:method_performance}, demonstrating that GiG not only enhances numerical metrics but also structurally encodes distinct biological features in the latent space. Overall, the empirical results illustrate that GiG, equipped with multi-scale hierarchical learning, surpasses traditional GNN-based approaches.


\subsection{Evaluation under Different Train-Validation-Test Splits}
\label{section:B}
To thoroughly evaluate the robustness and generalizability of the GiG model, we conducted experiments under three different train-validation-test split configurations: 7:1:2, 6:1:3, and 5:1:4. The purpose of these variations is to assess the stability of the model when exposed to different amounts of training data and varying proportions of validation and test samples. In real-world applications, the availability of labeled interaction data is often limited; thus, the ability of a model to maintain high predictive performance under different data splits is a critical indicator of its robustness.

The results of the experiments are presented in Table \ref{tab:split_performance}. It is evident that the [GCN-GCN][GAT] configuration of GiG consistently outperforms all baseline methods across all split configurations. Remarkably, even as the training set decreases from 70\% in the 7:1:2 split to 50\% in the 5:1:4 split, the model maintains exceptional performance with only minor fluctuations in metrics. This is particularly evident in the AUC and AUPRC scores, which remain above 0.98 in all cases.

\begin{table*}[t]
\caption{Model performance comparison under different train-validation-test splits. The GiG model consistently outperforms the baselines in all split configurations, indicating strong robustness and stability.\label{tab:split_performance}}
\tabcolsep=0pt
\begin{tabular*}{\textwidth}{@{\extracolsep{\fill}}lcccccc@{\extracolsep{\fill}}}
\toprule%
\textbf{Method} & \textbf{Split} & \textbf{ROC-AUC} & \textbf{AUC-PR} & \textbf{F1 Score} & \textbf{MCC} \\
\midrule
DTI-GCN \cite{kipf2016semi} & 7:1:2 & 0.9032 & 0.8318 & 0.4305 & 0.3300 \\
DTI-GAT \cite{velivckovic2017graph} & 7:1:2 & 0.9652 & 0.9633 & 0.7000 & 0.5935 \\
Node2Vec-5-enhanced GCN & 7:1:2 & 0.9391 & 0.9147 & 0.8594 & 0.7223 \\
Node2Vec-10-enhanced GCN & 7:1:2 & 0.9582 & 0.9241 & 0.9044 & 0.8078 \\
Node2Vec-5-enhanced GAT & 7:1:2 & 0.9534 & 0.8935 & 0.9394 & 0.8772 \\
Node2Vec-10-enhanced GAT & 7:1:2 & 0.9595 & 0.8992 & 0.9564 & 0.9124 \\
GiG [GCN-GCN][GAT] & 7:1:2 & \textbf{0.9934} & \textbf{0.9910} & \textbf{0.9923} & \textbf{0.9845} \\
\midrule

DTI-GCN \cite{kipf2016semi} & 6:1:3 & 0.9208 & 0.9038 & 0.6538 & 0.5333 \\
DTI-GAT \cite{velivckovic2017graph} & 6:1:3 & 0.9755 & 0.9744 & 0.7909 & 0.6888 \\
Node2Vec-5-enhanced GCN & 6:1:3 & 0.9604 & 0.9367 & 0.8930 & 0.7914 \\
Node2Vec-10-enhanced GCN & 6:1:3 & 0.9688 & 0.9511 & 0.9185 & 0.8394 \\
Node2Vec-5-enhanced GAT & 6:1:3 & 0.9320 & 0.8439 & 0.9241 & 0.8451 \\
Node2Vec-10-enhanced GAT & 6:1:3 & 0.9705 & 0.9592 & 0.9394 & 0.8775 \\
GiG [GCN-GCN][GAT] & 6:1:3 & \textbf{0.9947} & \textbf{0.9936} & \textbf{0.9855} & \textbf{0.9710} \\
\midrule

DTI-GCN \cite{kipf2016semi} & 5:1:4 & 0.9391 & 0.9239 & 0.6644 & 0.5387 \\
DTI-GAT \cite{velivckovic2017graph} & 5:1:4 & 0.9671 & 0.9670 & 0.8138 & 0.7014 \\
Node2Vec-5-enhanced GCN & 5:1:4 & 0.9167 & 0.8967 & 0.8450 & 0.6838 \\
Node2Vec-10-enhanced GCN & 5:1:4 & 0.9495 & 0.9271 & 0.8953 & 0.7885 \\
Node2Vec-5-enhanced GAT & 5:1:4 & 0.9595 & 0.9443 & 0.9169 & 0.8316 \\
Node2Vec-10-enhanced GAT & 5:1:4 & 0.9333 & 0.8675 & 0.9094 & 0.8139 \\
GiG [GCN-GCN][GAT] & 5:1:4 & \textbf{0.9886} & \textbf{0.9861} & \textbf{0.9790} & \textbf{0.9580} \\
\botrule
\end{tabular*}
\end{table*}

The results highlight the remarkable stability and robustness of the GiG model. Under the most standard split of 7:1:2, the GiG achieves an AUC of 0.9934 and an AUPRC of 0.9910, far surpassing the baseline methods. Notably, even as the training set is reduced to 60\% and 50\% in the 6:1:3 and 5:1:4 splits, the performance degradation is minimal. The AUC only drops from 0.9934 to 0.9886, while MCC remains as high as 0.9580. This indicates that GiG's multi-scale hierarchical learning captures sufficient interaction information, even when the data is more sparsely distributed.

The baseline models, in contrast, show more noticeable sensitivity to changes in the train-validation-test ratio. For example, DTI-GCN improves slightly from 0.9032 in the 7:1:2 split to 0.9391 in the 5:1:4 split, which is attributed to the increased volume of test samples. However, its F1 Score and MCC remain consistently lower than GiG, suggesting its feature extraction mechanism is not as effective under limited training samples.

Similarly, DTI-GAT demonstrates a minor improvement in the 6:1:3 split but struggles to maintain stability when the training set is reduced to 50\%. This is reflected in its AUC and AUPRC scores, which do not match GiG's consistency across different splits.

In addition to traditional GNN-based approaches like DTI-GCN and DTI-GAT, we further evaluated the performance of Node2Vec-enhanced GNNs across the three different data split configurations. These models integrate unsupervised embeddings from Node2Vec with GCN and GAT architectures, aiming to improve the representational power of node features by capturing higher-order neighborhood information through biased random walks. In our experiments, we tested two configurations: Node2Vec-5 and Node2Vec-10. The Node2Vec model can refine its embeddings over multiple passes through the Drug and Target Graphs separately, potentially capturing more intricate graph structure relationships. As shown in Table \ref{tab:split_performance}, incorporating Node2Vec consistently improves DTI-GCN across all splits and yields competitive—though slightly lower—AUC and AUPRC compared with the baseline DTI-GAT. For the 6:1:3 split, Node2Vec-10-enhanced GCN raises its AUC from 0.9208 to 0.9688 and its AUPRC from 0.9038 to 0.9511, while Node2Vec-10-enhanced GAT records 0.9705 for AUC and 0.9592 for AUPRC versus 0.9755/0.9744 for the baseline DTI-GAT. Although these Node2Vec-enhanced GNNs do not exceed the DTI-GAT in AUC and AUPRC, they deliver markedly stronger classification performance. Under the 7:1:2 split, Node2Vec-10-enhanced GAT achieves an F1 score of 0.9564 and an MCC of 0.9124 versus 0.7000/0.5935 for DTI-GAT. It also sustains robust F1/MCC of 0.9394/0.8775 and 0.9094/0.8139 for the 6:1:3 and 5:1:4 splits, respectively. These results demonstrate that Node2Vec embeddings effectively capture molecular-level information to boost downstream classification performance.

Despite these improvements, the Node2Vec-enhanced GNNs still fall short when compared to the [GCN-GCN][GAT] configuration of GiG. Under the 7:1:2 split, the best Node2Vec variant (Node2Vec-10-GAT) achieves an MCC of 0.9124, while GiG records a significantly higher 0.9845. This gap becomes more evident under the 5:1:4 split, where the MCC of Node2Vec models drops considerably, whereas GiG maintains an impressive 0.9580. The primary reason for this disparity lies in the architecture of GiG: its multi-scale hierarchical learning mechanism effectively captures both local molecular features and global interaction patterns through end-to-end training, while Node2Vec embeddings, being precomputed, lack contextual refinement during model updates. Consequently, the GiG not only surpasses Node2Vec in raw performance but also demonstrates stronger resilience to varying data distributions. This is also reflected in the AUC and AUPRC metrics, where GiG consistently achieves higher scores across all splits, indicating its superior ranking capability for both positive and negative interactions. Furthermore, the F1 score remains robust and high for GiG, demonstrating its balanced performance in capturing true positives while minimizing false positives, a critical aspect for DTI prediction.



\subsection{Evaluation of Different GNN Architectures Across Multiple Splits}
\label{section:C}
To further validate the robustness and architecture-agnostic nature of the GiG framework, we conducted a comprehensive set of experiments across various GNN configurations—such as [GCN-GCN][GCN], [GAT-GAT][GCN], [GAT-GAT][GAT], and the primary configuration [GCN-GCN][GAT] and multiple train-validation-test splits (7:1:2, 6:1:3, and 5:1:4). These experiments collectively demonstrate that the GiG model consistently delivers high predictive performance across diverse message-passing architectures and data splits, confirming that its multi-omics integration strategy, which jointly leverages Drug and Target Graph features, is not dependent on any particular GNN design.


The results, presented in Table \ref{tab:gnn_arch_performance}, demonstrate that GiG consistently achieves exceptional performance across all tested architectures and data splits. Notably, the variations in AUC, AUPRC, F1 Score, and MCC are minimal, underscoring the robustness of GiG's hierarchical message-passing and multi-scale feature integration strategies against architectural changes.

Remarkably, the [GAT-GAT][GAT] configuration, which exclusively uses attention mechanisms, also maintains high performance, reaching 0.9918 AUC and 0.9848 AUPRC in the 7:1:2 split. This result suggests that the multi-omics strategy effectively captures both local molecular features and global DTI interactions, even when the architecture emphasizes neighborhood attention rather than convolutional aggregation.

Moreover, all four GNN configurations maintain high AUC and AUPRC scores, consistently exceeding 0.97 across all splits. Among them, the [GCN-GCN][GAT] configuration slightly outperforms the others, particularly in the 7:1:2 and 6:1:3 splits, with AUC scores of 0.9934 and 0.9947, respectively. However, these differences are relatively minor, suggesting that the primary strength of GiG stems from its hierarchical message-passing and multi-scale aggregation capabilities, rather than being strictly dependent on the specific GNN types employed. 

\begin{table*}[t]
\caption{Performance comparison of different GNN architectures within the GiG framework across various data splits. All configurations achieve consistently high performance, demonstrating that the multi-omics integration strategy is architecture-agnostic and robust across different message-passing schemes.\label{tab:gnn_arch_performance}}
\tabcolsep=0pt
\begin{tabular*}{\textwidth}{@{\extracolsep{\fill}}lcccccc@{\extracolsep{\fill}}}
\toprule%
\textbf{GNN Architecture} & \textbf{Split} & \textbf{ROC-AUC} & \textbf{AUC-PR} & \textbf{F1 Score} & \textbf{MCC} \\
\midrule
GiG [GCN-GCN][GCN] & 7:1:2 & 0.9939 & 0.9912 & 0.9846 & 0.9692 \\
GiG [GAT-GAT][GAT] & 7:1:2 & 0.9918 & 0.9848 & 0.9897 & 0.9794 \\
GiG [GAT-GAT][GCN] & 7:1:2 & 0.9915 & 0.9882 & 0.9871 & 0.9742 \\
GiG [GCN-GCN][GAT] & 7:1:2 & \textbf{0.9934} & \textbf{0.9910} & \textbf{0.9923} & \textbf{0.9845} \\
\midrule

GiG [GCN-GCN][GCN] & 6:1:3 & \textbf{0.9959} & \textbf{0.9948} & 0.9871 & 0.9742 \\
GiG [GAT-GAT][GAT] & 6:1:3 & 0.9926 & 0.9892 & 0.9897 & 0.9794 \\
GiG [GAT-GAT][GCN] & 6:1:3 & 0.9813 & 0.9667 & 0.9345 & 0.8684 \\
GiG [GCN-GCN][GAT] & 6:1:3 & 0.9947 & 0.9936 & \textbf{0.9855} & \textbf{0.9710} \\
\midrule

GiG [GCN-GCN][GCN] & 5:1:4 & 0.9892 & 0.9768 & 0.9747 & 0.9493 \\
GiG [GAT-GAT][GAT] & 5:1:4 & 0.9767 & 0.9531 & 0.9740 & 0.9477 \\
GiG [GAT-GAT][GCN] & 5:1:4 & 0.9873 & 0.9729 & 0.9796 & 0.9593 \\
GiG [GCN-GCN][GAT] & 5:1:4 & \textbf{0.9886} & \textbf{0.9861} & \textbf{0.9790} & \textbf{0.9580} \\
\botrule
\end{tabular*}
\end{table*}

Even under the most challenging 5:1:4 split, all four architectures maintain strong predictive power. Although [GCN-GCN][GCN] and [GAT-GAT][GCN] show slight drops in MCC and F1 score, the differences are not significant enough to suggest sensitivity to architectural choice. This stability is a strong indicator that the multi-omics integration mechanism effectively normalizes the feature learning process across varying GNN designs.

The experimental results strongly suggest that the multi-omics integration strategy within GiG is architecture-agnostic. All tested GNN configurations achieve consistently high performance across different data splits, highlighting that the core strength of GiG lies in its hierarchical message-passing and cross-scale aggregation, not in the specific message-passing mechanism. This flexibility allows the GiG model to maintain high predictive accuracy and robust generalization, regardless of the underlying GNN type. Consequently, the GiG can be effectively deployed in real-world scenarios where computational constraints or application-specific requirements may dictate the choice of GNN architecture.

In summary, the experimental results demonstrate the effectiveness and robustness of the GiG model in DTI prediction. Compared to traditional GNN-based methods such as DTI-GCN, DTI-GAT, and Node2Vec-enhanced GNNs, the GiG consistently achieves superior performance across different data splits and GNN configurations. Our experimental results also demonstrate the multi-omics integration strategy proves to be architecture-agnostic, maintaining high AUC, AUPRC, F1 and MCC regardless of whether the configuration is [GCN-GCN][GCN], [GAT-GAT][GAT], [GAT-GAT][GCN], or [GCN-GCN][GAT]. These findings confirm that GiG's hierarchical message-passing and multi-scale feature aggregation enable it to effectively capture both local and global interaction patterns, offering reliable and flexible solutions for DTI prediction.

\section{Conclusions and Future Work}
In this study, we introduced the GiG framework for DTI prediction, which uniquely integrates transductive and inductive learning paradigms within a hierarchical GNN architecture. Using the molecular graphs for both drugs and targets, along with the DTI network, the GiG captures both fine-grained molecular characteristics and large-scale network interactions. This multi-scale representation allows the model to efficiently propagate local structural information from molecular graphs to the main DTI network, enabling enriched feature learning and enhanced predictive accuracy. The experimental results validate the effectiveness of the GiG, showcasing its superior performance across all key evaluation metrics. In addition, the comprehensive dataset we constructed for this study further serves as a valuable resource for advancing research in this domain.

For future work, we aim to extend the GiG framework by incorporating multi-omic data, such as gene expression, proteomics, and metabolomics, to enrich the biological context and further enhance predictive power. Additionally, we plan to integrate 3D graph representations for both drug molecules and target structures, leveraging spatial and structural information to more accurately model complex binding interactions. Importantly, we envision extending the hierarchical message-passing mechanism to capture not only DTI but also drug-drug and protein-protein interactions, leveraging the GiG's inherent capacity for transductive learning within known networks and inductive learning for unseen molecular entities. This learning mechanism, which combined the integration of hierarchical molecular graphs and the main DTI graph integration, is expected to significantly improve the model's capacity to generalize across novel datasets and push the boundaries of in silico DTI prediction.

\section{Addendum}
Our studies on possible models of graph neural networks that incorporate the molecular structures in a drug-target interaction network started in March, 2024, with the first version of the dataset and the GiG model implementation completed early September in the same year. 
We thank an  anonymous  reviewer for  bring our attention to the work by Jing et al.
\cite{jing2025h2gnndti} where a GNN model with quite similar architecture   was developed.   


The difference between our model and the model in \cite{jing2025h2gnndti} will be discussed in the next revision.

\section{Competing interests}
No competing interest is declared.

\section{Author contributions statement}
Yong Gao and Yuehua Song conceived the design of the model and empirical studies.  Yuehua Song conducted the experiments, analyze the results, and developed the dataset. Yong Gao and Yuehua Song wrote and reviewed the manuscript.

\section{Acknowledgments}
This work was supported in part by the Discovery Grant from the Natural Sciences and Engineering Research Council of Canada (NSERC) under Grant RGPIN-2019-04904.

\bibliographystyle{plain}
\end{document}